# Benchmarking language models and agents for scientific reasoning in evolving materials laboratories

**Authors**

Xuyu Dong[1,2,3], Yixuan Li[3], Jianguo Huang[3], Feng Chen[3], Yiyun Zhou[3], Bo An[3*], Zheng Liu[2*], Da Shu[1*]

**Affiliations**

[1] Shanghai Key Lab of Advanced High-temperature Materials and Precision Forming and State Key Lab of Metal Matrix Composites, School of Materials Science and Engineering, Shanghai Jiao Tong University, Shanghai 200240, China.

[2] School of Materials Science and Engineering, Nanyang Technological University, Singapore, Singapore.

[3] College of Computing and Data Science, Nanyang Technological University, Singapore, Singapore.

* Corresponding authors: Prof. Bo An, Email: boan@ntu.edu.sg; Prof. Zheng Liu, Email: z.liu@ntu.edu.sg; Prof. Da Shu, Email: dshu@sjtu.edu.cn.

**Abstract**

Large language models and the multi-agent systems built around them could accelerate materials discovery by connecting scientific knowledge with experimental decisions. Existing benchmarks establish capabilities in domain knowledge, multimodal interpretation and scientific tool use. As language models increasingly support materials research workflows, their ability to distinguish closely related material states and reason accurately with quantitative evidence remains insufficiently evaluated. These abilities are essential for comparing material responses, assessing mechanisms and maintaining reliable judgements as experimental evidence accumulates. We introduce MatScope to evaluate three connected capabilities. M1 tests unaided, condition-specific reasoning through 200 tasks spanning ten research functions, using aluminium and high-entropy alloys as representative material systems. M2 tests agent-assisted transfer of related knowledge to unresolved materials questions. MatEvoBench-20 tests initial forecasts across scientific dependency chains and revisions of still-hidden responses as observations accumulate. Across eleven model deployments, including GPT-6 Astra and Claude Fable 5.1, given-input calculations and many physical explanations are handled well. Fine-grained response estimates, comparisons between similar states and quantitatively supported choices remain substantially weaker. Related literature improves selected estimates, while condition and baseline mismatches limit transfer. Progressive observations calibrate some downstream responses and destabilize earlier accurate predictions. MatScope establishes measurable targets for developing reliable scientific reasoning in future self-evolving materials laboratories and materials-aware world models.

## Introduction

Large language models (LLMs) and their multi-agent architectures are creating new routes to materials discovery. They connect scientific literature, computational models and experimental interfaces through flexible reasoning and tool coordination. Systems such as AtomAgents integrate language-based planning with physics simulations for alloy design, while Coscientist demonstrates language-model-driven experimental chemistry [1,2]. These advances make autonomous and AI-assisted laboratories a practical research direction. Their success also raises a central evaluation question: which scientific judgements can the language model reliably contribute to a workflow whose outcome depends jointly on models, retrieved knowledge, optimization and experimental control[3,4] ?

Existing benchmarks provide essential foundations. MaScQA and MatSciBench evaluate materials knowledge, numerical problem solving and structured scientific reasoning [5,6]. MaCBench extends assessment to multimodal extraction, experimental understanding and interpretation[7]. Strong performance on established knowledge tasks documents genuine progress: a common-protocol MaScQA evaluation reported 60.8% overall accuracy for GPT-4 and approximately 84% for GPT-4o and Claude-3.5-Sonnet 5. LightAlloy-Bench evaluates composition–process–property reasoning through mechanical-property selection and identifies numerical and mechanistic limitations[8]. These studies motivate a broader assessment of how language models interpret material states, evaluate quantitative evidence and maintain scientific judgements as research conditions and available information evolve [9-15].

Agent benchmarks address complementary requirements. AstaBench and SciAgentGym evaluate research-tool access, execution and multi-step coordination [16,17]. AILA and AFMBench assess real AFM operations, including planning, instrument control and analysis within an atomic-force-microscopy workflow [18]. This instrument-centred evaluation establishes important operational capabilities. Materials research also requires state-specific interpretation, comparison and inference across processing branches and structural scales. Evaluating these connected judgements links tool execution to the scientific understanding needed throughout an investigation [19-23].

The distinction matters because material responses are path-dependent and multiscale, making quantitative understanding essential for reliable reasoning throughout an extended laboratory investigation. Atomic-scale chemistry and crystallography constrain nanometre-scale precipitates and interfaces. Micrometre-scale grains, segregation and defects influence specimen-scale strength, ductility and degradation. Several mechanisms act concurrently, and their relative contributions change with processing [24-26]. Particle sizes and spacings, phase fractions and defect densities distinguish material states and constrain the contributions of these mechanisms. An incorrect phase fraction can misrepresent transformation progress, while an incorrect particle spacing can distort a strengthening calculation. Such errors affect state identification, mechanism assessment and comparisons between specimens. They can also enter subsequent reasoning as incorrect premises when later observations are interpreted. These requirements also apply to values supplied by simulations, numerical tools and experiments. Reliable materials judgement therefore requires the correct state, physical quantity, magnitude and comparison to be maintained throughout the investigation[24-26].

An evolving investigation requires three connected forms of competence. First, a model must apply its existing knowledge to the stated conditions and predict unknown responses. Its deployed weights remain fixed during the evaluation. Second, unresolved questions require the acquisition and adaptation of related knowledge. A useful analogy must preserve the relevant composition, processing regime and measurement definition. Third, observations from the current experiment must constrain predictions that remain untested. Effective adaptation combines accurate initial expectations, selective revision and sustained accuracy.

Here we introduce MatScope to evaluate the materials reasoning required of LLMs and agents in self-evolving laboratories, using alloy systems as representative cases within a generalizable evaluation framework. M1 tests unaided, node-level judgement through 200 alloy tasks spanning ten research functions. M2 tests whether agents can improve unresolved M1 node-level judgements by acquiring and applying knowledge from related studies. MatEvoBench-20 (ME20) evaluates these types of judgement across ninety nodes in twenty multistage research trajectories. Most nodes are connected through material evolution, shared states or cross-scale relationships, allowing evaluation of how successive observations affect reasoning across an investigation. In the 11 front-line model evaluations, the success rate of the M1 task ranged from 13.5% to 19.0%. The questions with universal material explanations and clear input calculations had high success rates (92.38–99.05%), but performed poorly in making precise and granular inferences (22.74–31.21%). On twenty selected M2 failure cases, GPT-6 Astra and Claude Fable 5.1 gain 1.60 and 4.22 percentage points, respectively. Across 48 shared ME20 endpoints, numerical rubric scores rise from 32.00% to 37.50% for Astra and from 21.42% to 37.67% for Fable between initial and last-hidden assessments. However, only four of seven initially top-band endpoints for Astra and one of three for Fable retain that accuracy throughout updating. Together, these evaluations distinguish limitations in initial materials understanding, transfer of external knowledge and maintenance of accurate judgements as evidence accumulates. This indicates that existing large language models still face difficulties in making continuous and in-depth correct inferences in the self-evolving material laboratory scenarios. MatScope provides measurable targets for scientific intelligence in future self-evolving materials laboratories and for materials-aware world models.

## Results

### A capability framework for evolving materials laboratories

MatScope evaluates whether LLMs and agents can maintain accurate materials understanding from local judgements to linked research trajectories (Fig. 1a–c). Laboratory reasoning includes interpreting characterization, comparing specimens, assessing mechanisms and revising plans. Each task therefore specifies a material, processing branch, specimen state and measurement object. Particle spacing, grain size and phase fraction remain distinct variables. Parallel heat treatments retain separate histories. The same requirements apply to values supplied by simulations, numerical tools or experiments. High scores require the requested quantities and relationships to be correct for the specified material conditions.

M1 and M2 examine local scientific judgements, while MatEvoBench-20 (ME20) examines their connection across a research pathway. M1 evaluates unaided interpretation, comparison and estimation for specified material states. M2 provides literature-search, reading and calculation tools for selected unresolved M1 requirements. The construction-source publications are excluded from retrieval. The agent must assess which relationships and response scales can inform the target problem. ME20 places comparable operations in independent research cases. Registered relationships connect processing branches, shared states and cross-scale measurements. Models first predict all requested hidden targets, then revise those still hidden after each observation. For example, grain size and matrix solute content can inform a strength estimate before tensile results are released.

Task construction draws on an inventory of 28,872 deduplicated literature works, with evaluation cases primarily covering aluminium alloys and high-entropy alloys. Bibliographic metadata, full-text passages and structured composition–process–observation–property records describe complementary aspects of the corpus. Embeddings and Uniform Manifold Approximation and Projection (UMAP) show thematic distribution and the locations of task sources (Fig. 1d) [27]. Source selection requires identifiable samples, explicit conditions and supported quantitative or relational targets. Structured extractions locate candidate records. Original measurements and source-supported relationships establish the references. Coverage analyses retain material families, processing routes, characterization methods and properties separately. These summaries document sampling concentration and gaps within the collected corpus (Supplementary Notes 1–2 and Supplementary Table 1).

M1 contains 200 tasks and 1,124 targets across 96 source-connected groups. Its ten-function distribution is reported in Table 1 and Supplementary Fig. 1. M2 examines twenty shared-failure cases. ME20 comprises twenty investigations with ninety nodes. English questions specify the required conditions and response definitions. Reference outcomes and scoring rules remain separate from the model input. Target-specific rubrics assign graded credit for numerical proximity and precise scientific relationships. Numerical accuracy, ordering, mechanism scope and feasibility retain separate checks. Repeated responses measure numerical stability relative to the declared scoring resolution. Experimental variability, model variability and evaluator discrepancies are analysed separately. These measures distinguish partial understanding, accurate quantitative discrimination and completion of the required scientific goals.

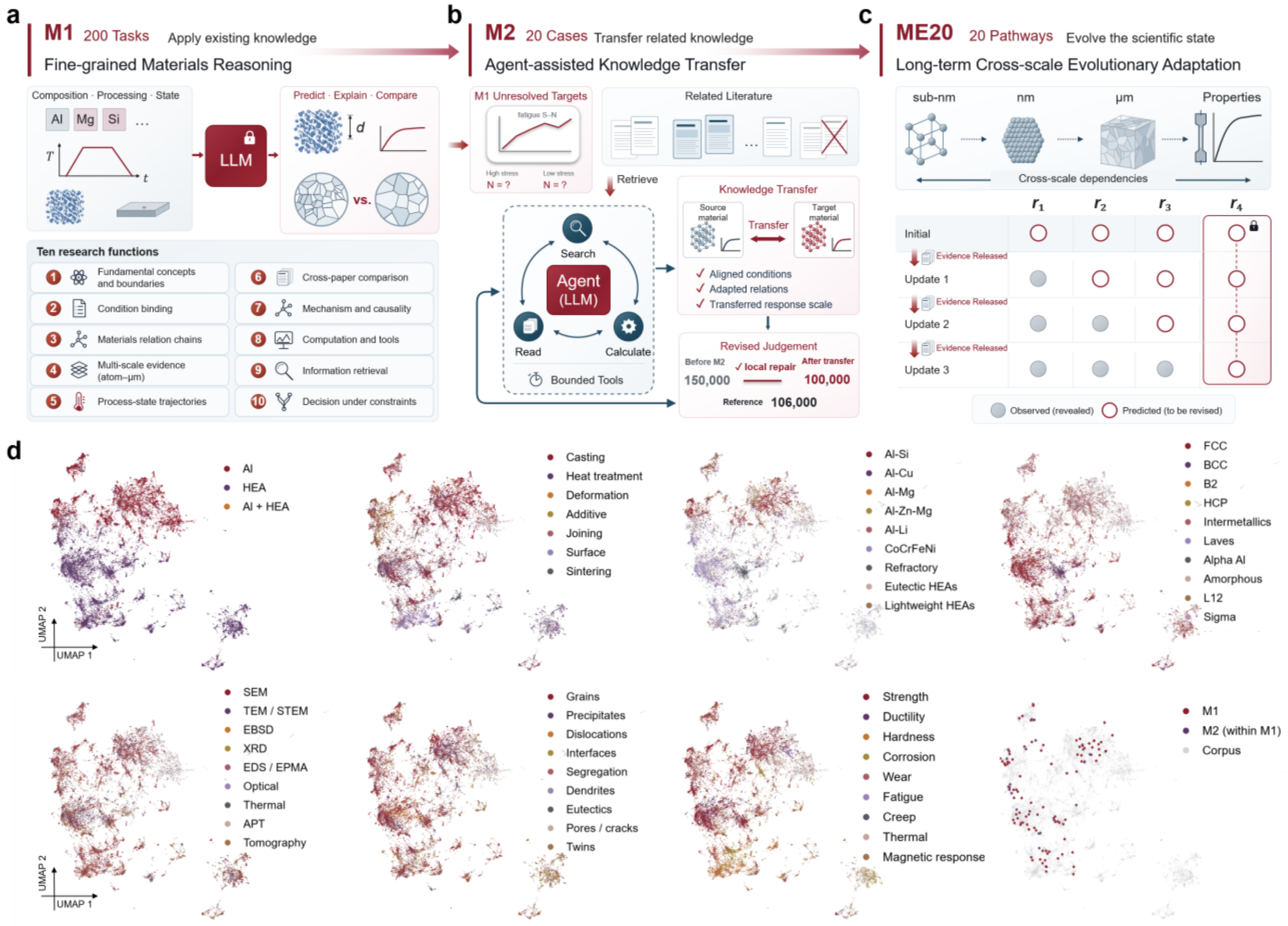

**Fig. 1 | Materials judgement from local tasks to linked research pathways. a,** M1 evaluates unaided, condition-specific judgement across ten materials functions. **b,** M2 evaluates agent-assisted knowledge transfer on selected unresolved M1 requirements. The agent uses bounded literature-search, reading and calculation tools, with construction-source publications excluded. **c,** ME20 evaluates initial forecasts across linked material nodes and revisions of still-hidden responses as observations arrive. The matrix distinguishes observed nodes from predictions whose reference outcomes remain hidden. **d,** Joint UMAP views coloured by alloy domain, processing route, alloy system and phase labels. The same coordinates show characterization, microstructure and property labels, together with the construction publications supporting M1 and the selected M2 cases. Each point represents one deduplicated literature work. Of 28,872 works, 28,859 have embeddings and

28,555 fall within the displayed window. All UMAP views share one fit using 20 neighbours, a minimum distance of 0.05, cosine distance and seed 2608. UMAP coordinates are dimensionless. These views describe corpus composition and task-source coverage.

**M1: Materials knowledge for quantitative experimental decisions**

M1 examines how far pretrained knowledge can support the operations required for a materials decision (Table 1, Fig. 2). Its ten functions preserve condition binding, quantitative comparison, cross-scale interpretation, processing histories and feasible route selection. Eleven deployments obtain mean rubric scores of 58.59–68.75% and complete task success of 13.5–19.0% across 200 tasks (Fig. 2a). Quantitative comparison is a shared weakness across all eleven deployments. Scores range from 40.47% to 53.46%, compared with 72.29–92.67% for limits of inference. The models are substantially better at identifying the scope of a conclusion than at determining the magnitude, direction and baseline of material differences (Table 1; Fig. 2b). Figures 2b,c compare the ten research functions and four numerical-observable groups across all eleven deployments. Supplementary Fig. 1 shows the material-domain composition of each function.

The $Fe_{45}Co_{35}Cr_{10}V_{10}$ case concerns an 80% cold-rolled alloy annealed at 750 °C for 300 s. Astra predicts 100% FCC mapped area and 50% recrystallization, Fable predicts 97% and 80%, and GPT-6 Sol predicts 95% and 25%, against respective references of 99.9% and 57.8%. All three models identify a nearly complete FCC state. Their recrystallization estimates deviate from the reference by −7.8, +22.2 and −32.8 percentage points for Astra, Fable and Sol, respectively. Phase reversion is therefore captured more accurately than the extent of recrystallization. The models only partially resolve the concurrent transformations within this annealed state. The strength–ductility decision in the die-cast Al–Si–Mg–Mn case and the route-dependent copper-addition comparison further distinguish quantitative inputs from the corresponding decision or transfer relationship. These cases separate accuracy in material quantities from accuracy in the associated relations or decisions (Fig.

2d). Answer-generation cost and score form a separate deployment comparison (Fig. 2e). The observed frontier is conditional on these configurations and excludes grading cost. Unit and ordering repairs use the same saved final and programmatic rule across deployments. The CS–CDS zinc-dispersion case's recorded semantic-relation concern remains a separate evaluator limitation (Supplementary Note 3, Supplementary Figs. 2–8, Supplementary Tables 2–3) [26,28-30].

Models execute supplied quantitative relations far more reliably than they estimate responses for specific material states. Across 200 tasks, the 49 given-input calculation targets receive weighted normalized scores of 92.38–99.05%, compared with 22.74–31.21% for the 476 conditional numerical predictions. The calculation targets test formulas, units and arithmetic. The prediction targets test whether the model can supply quantities appropriate to the specified material state (Fig. 3a).

Many estimates are too inaccurate to distinguish nearby material states quantitatively. Across 75 common positive-valued targets, median factor errors range from 1.59 to 2.12 (Fig. 3b). A factor error above two places the estimate below half or above twice the reference. Such estimates fail to establish the requested material magnitude at the precision needed for close comparisons. The broader conditional-prediction analysis also records 26 errors of at least tenfold for Astra and 23 for Fable among their positive submissions (Supplementary Fig. 10). The weakness therefore extends from fine precision to gross errors in material magnitude.

Numerical estimates are also insufficiently repeatable at the resolution required for effective materials judgement. Across nineteen comparable triplets per deployment, median spans reach 4.57 and 5.00 grade-five error thresholds for Astra and Opus 5.5, respectively (Fig. 3c). Only seven Astra triplets and four Opus triplets remain within one such threshold. Derived quantities show larger normalized spans than direct responses, with medians of 20.00 versus 1.67 for Astra and 10.47 versus 4.33 for Opus (Supplementary Fig. 14). Repeated answers thus provide different numerical premises under unchanged experimental conditions. This variability weakens the reproducibility of subsequent state comparisons and scientific reasoning.

In the TiC-modified Al7068 case, the source gives a mean diameter of 180 nm for constituent TiC particles within agglomerates and a mean inter-agglomerate spacing of 6 μm. Astra predicts 60 nm and 0.5 μm. Fable predicts 200 nm and 2 μm. Their answers distinguish relevant particle-strengthening and grain-pinning roles, yet underestimate spacing twelve-fold and three-fold. Mechanism recognition exceeds the models' ability to quantify the microstructure to which the mechanism applies. The answers identify relevant strengthening and pinning roles, yet misrepresent the particle spacing needed for quantitative analysis (Fig. 3d). The Zn-dispersion case exposes an error in the processing comparison itself. After 12 h at 450 °C, the reference standard deviation is 40.3% lower in CDS than in CS. Submissions above the equality line reverse this ordering and identify the wrong route as having lower intragranular Zn dispersion (Fig. 3e). State ratios for $MgZn_2$ diameter and spacing further distinguish preservation of a heat-treatment response from recovery of its absolute scale (Fig. 3f). Together, these cases show that plausible material explanations can coexist with incorrect magnitudes and reversed comparisons. Those errors leave the specific material state and processing response poorly resolved [31,32].

Matching an average magnitude can conceal a completely wrong material comparison. DeepSeek Pro predicts a −40-K temperature offset for all four compositions, against references of +61, +55, +36 and +5 K. Its mean absolute offset of 40 K is close to the 39.25-K reference average. However, every sign is reversed and the composition dependence is lost. The mean absolute error is 79.25 K. The answer therefore fails both the direction and the composition sensitivity of the required correction. Across 136 common tasks containing prediction and mechanism targets, only 1–4 tasks per deployment satisfy both requirements. Every prediction must reach grade 5, and every mechanism target must reach at least grade 4. The central weakness is combining a relevant mechanism explanation with accurate quantities for the same material state.

The overall scores of GPT-6 Sol and Astra are similar, but there are significant differences in their fine-grained material capabilities. GPT-6 Sol and Astra obtained 67.1955% and 67.6616%, respectively. The paired difference (Sol minus Astra) was -0.4660 percentage points (source-group bootstrap 95% interval, -2.24 to 1.29). Sol scored higher on 69 tasks, Astra on 70, and 61 received equal scores. Their mean absolute task-level difference was 8.87 points (Fig. 3g,h). Sol's mean advantage was 8.98 points in joint-source tasks and 2.42 points in model-calculation tasks, whereas Astra's advantage was 10.43 points in process–state relations. The joint-source stratum contains only eight tasks and six source groups. These post hoc descriptive contrasts identify different bottlenecks within the specified task set. The differences describe the functions and materials in this task set. Better performance requires state-specific quantitative responses and coherent relations, with specific quantitative detail.

**Table 1 | Materials functions assessed in M1**

Scores are the lowest and highest adopted mean task scores across the eleven deployments. They describe model variation within each function. Statistical confidence intervals are reported separately.

| Function | Tasks | Scientific operation and experimental relevance | Score range (%) |
|---|---|---|---|
| Foundational anchors | 2 | Apply basic quantities, units and physical relations. | 100.00 |
| Nearby-condition binding | 28 | Assign a response to the correct composition, treatment and state. | 59.80–70.26 |
| Quantitative comparison | 25 | Preserve magnitude, direction, ordering and comparison baseline. | 40.47–53.46 |
| Cross-scale interpretation | 37 | Link atomic/sub-nm identity, nm particles and interfaces, μm grains and defects, and specimen-scale properties. | 58.38–73.59 |

| Function | Tasks | Scientific operation and experimental relevance | Score range (%) |
| --- | --- | --- | --- |
| Processing–state relations | 24 | Track branches, sequence and intermediate states through processing. | 54.95–69.30 |
| Cross-source integration | 8 | Combine complementary, condition-compatible findings. | 63.60–77.64 |
| Constrained materials decisions | 27 | Select candidates that satisfy simultaneous response requirements. | 49.46–62.58 |
| Models and computational applicability | 25 | Apply calculations and identify the conditions supporting model use. | 67.15–75.50 |
| Bounded transfer | 15 | Determine which relationships survive a change of material or route. | 64.90–76.25 |
| Limits of inference | 9 | Limit conclusions to the measurements and comparisons that support them. | 72.29–92.67 |

All task-score and target-grade comparisons use 200 tasks. Supplementary Note 1 specifies the reporting convention.

Length scales describe scientific entities across the task set. Characterization techniques overlap in their spatial sensitivity. Task-specific settings determine the relevant scale. Function, target type and information setting form separate analysis axes.

Matched material-failure contrasts and saved numerical-repeatability diagnostics are detailed in the Supplementary Information (Supplementary Notes 4–5, Supplementary Figs. 9–14, Supplementary Tables 4–6).

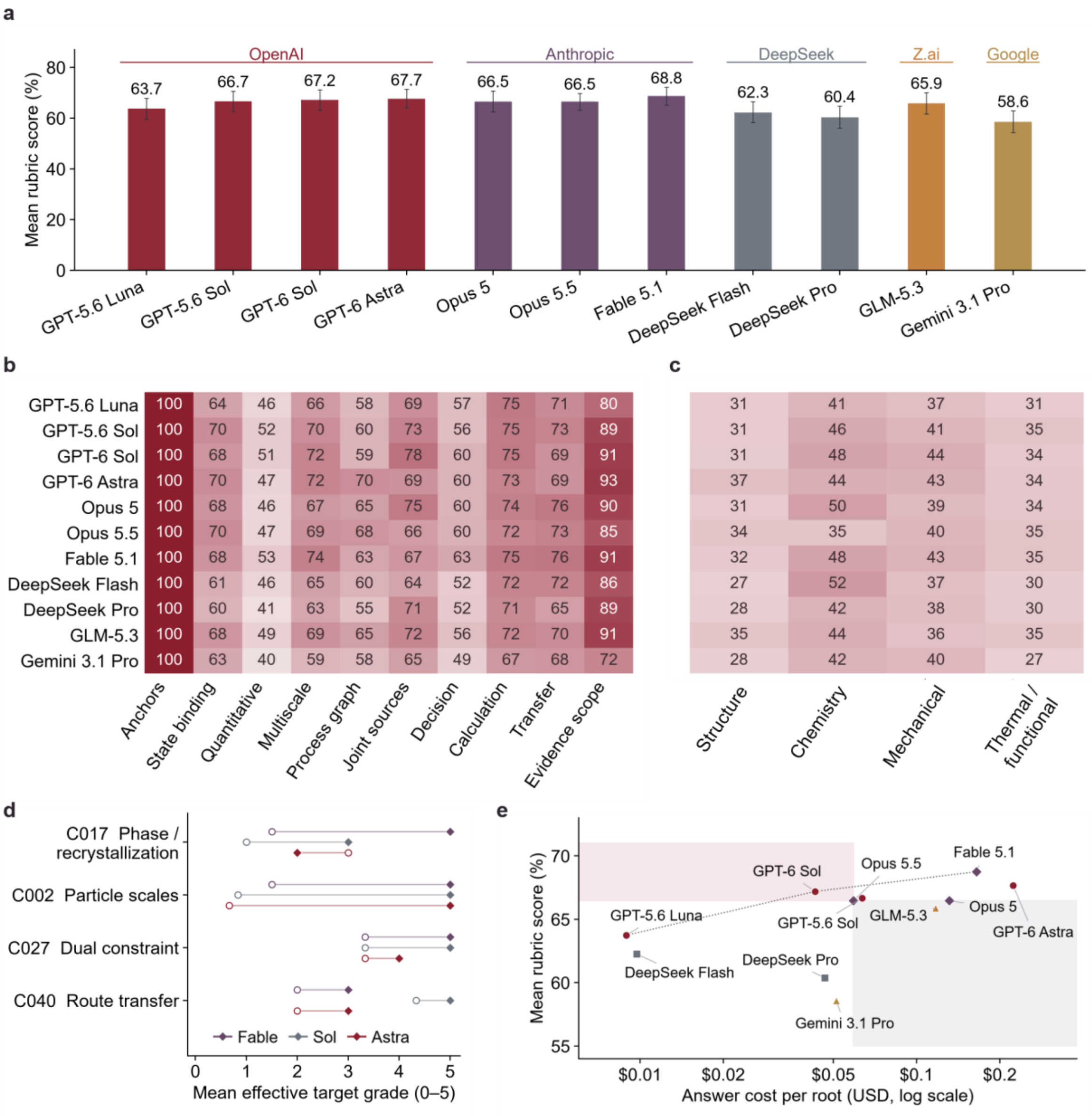

**Fig. 2 | Functional strengths and failures in materials reasoning. a,** Mean weighted task score on 200 roots, grouped by model family. Each bar is a deployment. Error bars are 95% intervals from 2,000 resamples of source-connected groups. **b,** Rows are the eleven deployments and columns the ten materials functions. Cells give task-equal mean scores. **c,** Numerical grades grouped by structure and scale, local chemistry, mechanical response, and thermal or functional response, with within-task weight normalization. **d,** Four material cases contrast numerical grades (open circles) and relation or decision grades (filled diamonds) on the 0–5 scale. Lines connect the two components of one deployment. The cases address phase reversion, particle strengthening, joint strength–ductility choice and copper-addition trends. **e,** Answer-generation cost per task on a

logarithmic horizontal axis versus mean task score. Each point is a deployment. The dotted line connects the observed Pareto frontier, and the background division uses cohort medians.

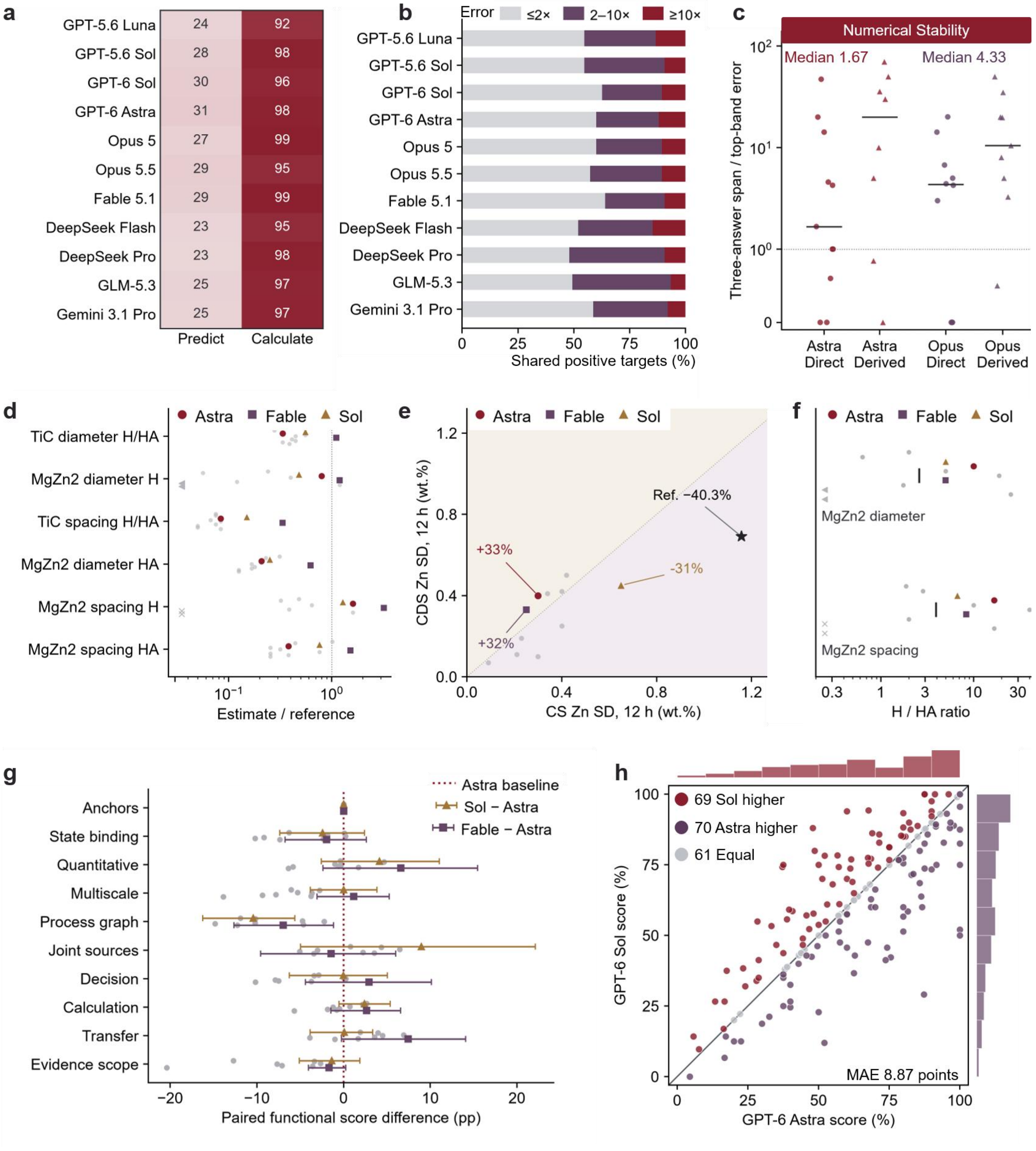

**Fig. 3 | Quantitative errors, repeatability limits and functional differences. a,** Weighted scores for 476 conditional numerical predictions and 49 given-input calculations on 200 roots. Rows denote deployments. **b,** Factor-error distributions for 75 common positive-valued targets. Factor error is the larger of the estimate-to-reference and reference-to-estimate ratios. Segment lengths show the fraction of targets within each category. Errors above two correspond to estimates below

half or above twice the reference. **c,** Repeatability under unchanged task conditions. Each point represents the span of three independent numerical submissions, normalized by the target's grade-five error threshold. Absolute-error targets use the linear range. Logarithmic-error targets use the log-ratio range. Direct and derived quantities are shown separately. Horizontal bars mark medians. d, Particle dimensions divided by their experimental references on a logarithmic axis. The TiC diameter refers to constituent particles within agglomerates, whereas the TiC spacing reference is the distance between agglomerates. A ratio of 1 matches the reference. Symbols identify deployments. Astra submits 100 nm and Fable 150 nm for a 125-nm $MgZn_2$ diameter, both receiving 3/5. e, Paired estimates of intragranular Zn standard deviation after 12 h at 450 °C. The horizontal axis gives CS and the vertical axis CDS, in wt.%. Each point contains one deployment's two estimates. The star at (1.157, 0.691) gives the experimental reference from forty filtered intragranular EDS points per state. The diagonal means equal dispersion. Points below it predict lower dispersion in CDS. f, H-to-HA ratios of $MgZn_2$ diameter and spacing, with reference ratios 2.60 and 3.94. d–f include all eleven deployments. g, Function-specific mean score differences relative to Astra. Highlighted contrasts are GPT-6 Sol minus Astra and Fable 5.1 minus Astra. Positive values favour the comparison deployment. Error bars show 95% intervals from paired source-group bootstrap resampling. h, Each point is one of the 200 roots, comparing Astra on the horizontal axis and GPT-6 Sol on the vertical axis.

**M2: Agent-assisted knowledge transfer for unresolved materials problems**

M2 evaluates how agents mobilize prior research when their unaided materials judgements remain incomplete (Fig. 4). Its scientific demand is to identify compatible evidence and translate it into a constraint on the current problem. Related studies may supply a mechanism, scaling relation or response range. Successful transfer requires preserving the material state and experimental comparison to which that information applies. Astra and Fable each answer twenty cases using a bounded literature-and-calculation harness. The paired analysis retains 64 shared-failure subquestions and 66 targets, including 59 numerical targets (Supplementary Note 6, Supplementary Table 7) [33,34].

Agent-assisted knowledge transfer produces limited and uneven improvements across the selected unresolved materials problems. Astra changes from 17.82% to 19.42% (Fig. 4a), a gain of 1.60 percentage points with a source-group 95% interval of −3.60 to 6.75. Fable changes from 15.19% to 19.41%, gaining 4.22 points with an interval of 1.01 to 7.73. Complete success is zero for both models on the selected requirements. Raw numerical errors decrease on 14 targets for Astra and 26 for Fable, and increase on 17 and 23. Five and fourteen improvements remain within the same scoring band. Continuous errors reveal useful movement towards the required response, while the bands retain its unresolved precision requirements (Table 2, Supplementary Table 8).

**Table 2 | Key outcomes in M2 and ME20**

M2: 20 cases, 64 selected subquestions and 66 targets. 59 numerical targets are comparable per model. Scores use the same selected requirements in M1 and M2. Complete success requires all selected critical requirements.

ME20: the same 48 numerical endpoints from 20 cases and 20 source groups are compared before any observations and at each target's last hidden checkpoint. Values are case-equal numerical rubric scores. Error counts compare absolute errors within each target's unit. Retention requires remaining in the top band at every hidden checkpoint.

| Outcome | GPT-6 Astra | Claude Fable 5.1 |
|---|---|---|
| M2 paired failed-subquestion analysis | | |
| Unaided M1 score (%) | 17.82 | 15.19 |
| M2 score (%) | 19.42 | 19.41 |
| Gain, percentage points (95% CI) | +1.60 (-3.60 to 6.75) | +4.22 (1.01 to 7.73) |
| Complete success / cases | 0 / 20 | 0 / 20 |
| Error closer / farther / unchanged | 14 / 17 / 28 (59) | 26 / 23 / 10 (59) |
| ME20 common hidden numerical endpoints | | |

| Outcome | GPT-6 Astra | Claude Fable 5.1 |
|---|---|---|
| Initial forecast score (%) | 32.00 | 21.42 |
| Last-hidden forecast score (%) | 37.50 | 37.67 |
| Gain, percentage points (95% CI) | +5.50 (-6.67 to 19.50) | +16.25 (3.67 to 29.67) |
| Top-band endpoints, initial → last | 7 / 48 → 8 / 48 | 3 / 48 → 6 / 48 |
| Error closer / farther / unchanged | 16 / 9 / 23 (48) | 21 / 14 / 13 (48) |
| Initially top-band endpoints retained throughout | 4 / 7 | 1 / 3 |

The tool-use accounting separates actions, returned passages, applicability review and explicit use chains (Fig. 4b). Higher scoped target grades occurred on nine of 66 Astra targets and twelve of 66 Fable targets. 47 remained unchanged for each deployment (Fig. 4c). Across the 40 episodes, Astra and Fable received means of 8.2 and 8.1 distinct source handles, respectively. Astra made 29 additional passage-reading calls and Fable four, despite similar search breadth (Fig. 4d). Issued-search outcomes also distinguish returned evidence, no-passage responses, access restrictions and action-budget limits (Fig. 4e). These records characterize evidence acquisition and reading activity, while the target-level outcomes show whether the acquired information improves materials judgements (Supplementary Table 9, Supplementary Figs. 15–19).

Recorded interactions show why access and application can diverge. Fable 5.1 uses related fatigue information to revise a high-stress life estimate from 150,000 to 100,000 cycles, approaching the 106,000-cycle reference, while its lower-stress estimate remains inaccurate (Fig. 4f). This case illustrates how related knowledge can improve one response without resolving other conditions within the same material problem. Another answer transfers undercooling values from LM25 to LM6 despite differences in alloy composition and cooling conditions (Fig. 4g). These differences affect the applicability of the transferred values to the target alloy. In an ageing problem, Fable acquires a correct 386 MPa strength but combines an alloy-modification comparison with the baseline for within-alloy ageing. The required contrast is 299 to 386 MPa for the same alloy (Fig. 4h). The model obtains the correct strength but uses an incorrect comparison baseline, showing that accurate evidence retrieval does not necessarily lead to a valid materials comparison. Across fourteen paired numerical targets per model with reviewed evidence links, the continuous-error analysis shows both improvements and deteriorations after knowledge transfer (Fig. 4i). These targets have resolved material-condition matches, allowing the analysis to examine quantitative changes where the applicability of the retrieved evidence has been reviewed. Together, the cases highlight the challenge of calibrating related knowledge to the target material system [35].

Functional outcomes reinforce this interpretation. Fable improves most on processing–state relations and quantitative comparison, by 8.00 and 7.62 points, while finite-decision cases decline for both models. Several individually plausible material estimates still fail to satisfy the combined requirements of a constrained decision. Removing the largest-gain source group leaves Fable with a 3.04-point gain and Astra with 0.11, revealing different sensitivity to case composition. Construction-source exclusion, passage permissions and input scope define this experiment. It measures transfer through the complete agent-and-retrieval configuration. These node-level capabilities also matter in evolving materials investigations, where transferred knowledge may inform subsequent state interpretation, mechanism assessment and evidence-driven revision. Generation cost and the scoped outcomes are reported together (Supplementary Fig. 20).

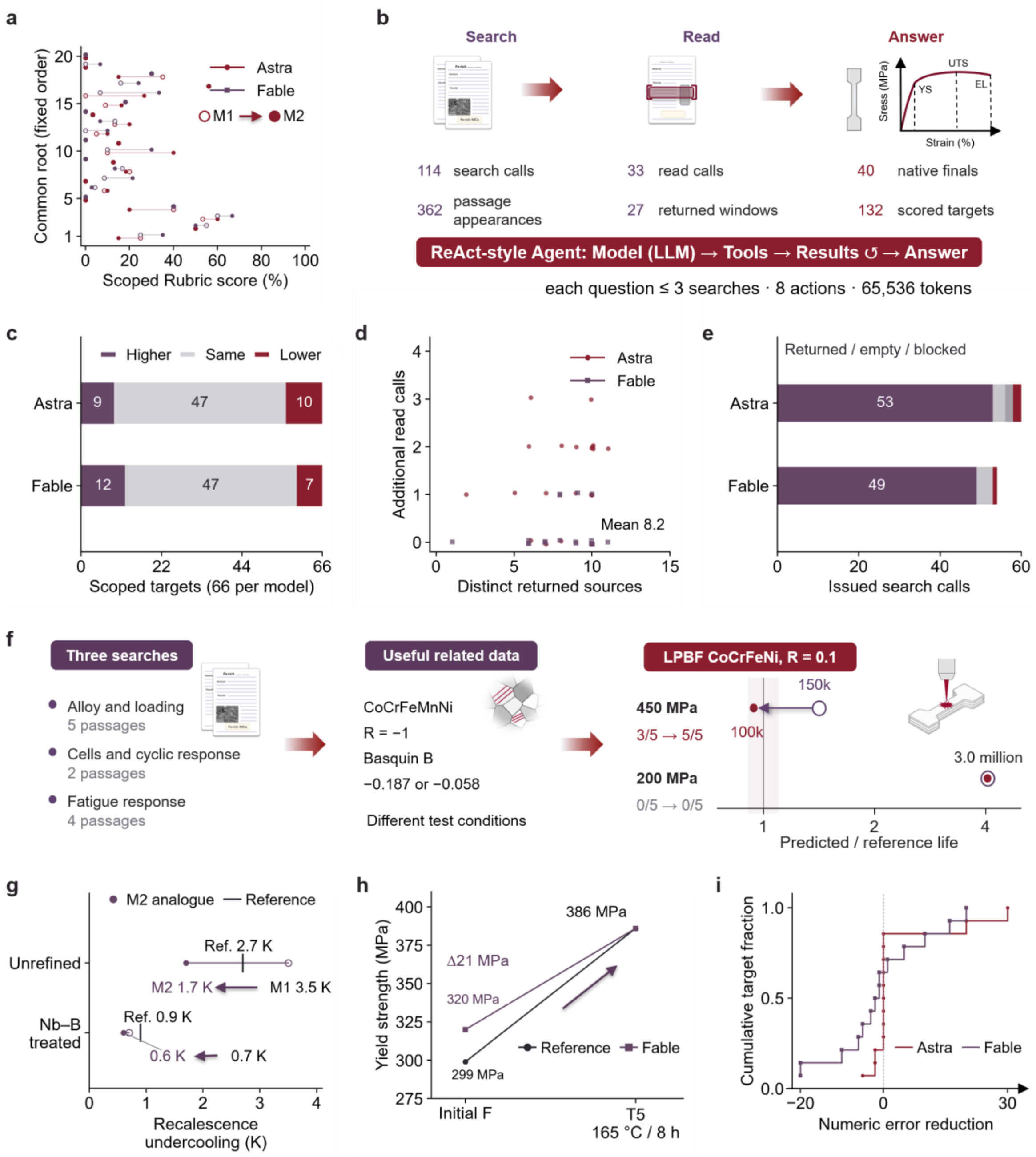

**Fig. 4 | Source-excluded knowledge transfer and material-condition binding. a,** M1 and M2 scores on identical selected requirements for twenty material problems. Open points are unaided M1 and filled points M2 after search. b, Counts of issued tool actions, returned text and saved outputs across forty episodes. Searches, reads, passage appearances and target grades retain separate counting units. c, Numbers of higher, unchanged and lower grades across 66 targets per deployment. d, Each point is an episode. The horizontal axis counts distinct document handles returned, and the vertical axis additional reading calls. Small jitter separates coincident integer coordinates. e, Search calls grouped by actual return status. f, Fable's three successive queries return 5, 2 and 4 passages.

The quantitative comparison shows initial (open purple) and post-search (filled red) fatigue-life estimates divided by the target reference. The vertical line at 1 denotes exact agreement, and the shaded band is the highest-grade interval. At 450 MPa, the estimate changes from 150,000 to 100,000 cycles against 106,000 cycles, increasing the numerical grade from 3/5 to 5/5. At 200 MPa, it remains 3,000,000 cycles against 740,000 cycles, receiving 0/5. Related-alloy measurements use a different stress ratio. g, LM25 and A356 analogues compared with the target LM6 recalescence undercooling under different composition and cooling conditions. h, Yield-strength comparisons from the initial F state to T5. The reference connects 299 and 386 MPa, whereas the model connects 320 and 386 MPa. The reference increase is 29.10%. i, Cumulative numerical-error reduction for fourteen condition-matched targets per deployment. The horizontal axis is M1 error minus M2 error, normalized by each target's highest-grade error allowance. Positive values denote estimates closer to the reference.

**MatEvoBench-20: Cross-scale predictive adaptation along experimental pathways**

ME20 tests whether models can maintain accurate materials judgements as evidence accumulates across linked states and measurements (Fig. 1c). The initial submission forecasts the full sequence of required responses. Later submissions revisit the same hidden targets as observations arrive. Scientific dependencies connect atomic structure, precipitates, grains, defects and macroscopic properties across processing or loading histories. Each prediction is committed before its verifying observation is released. This design assesses the prospective value of evidence and the persistence of accurate judgement. Twenty nodes represent later physical states, and seventy represent unrevealed measurements of an existing state. The ninety nodes cover grains and features, phases and particles, local chemistry, defects and damage, interfaces and layers, and material responses (Fig. 5a). The T6 case holds the material state fixed while grain, solute, precipitate and tensile records are revealed in sequence (Fig. 5b). It tests whether measured structure constrains the model's judgement of properties that remain hidden (Supplementary Note 7, Supplementary Table 10, Supplementary Fig. 21)[36].

Initial forecasts misrepresent both the magnitude and the pattern of material evolution (Fig. 5c,d).

Both models underestimate dislocation accumulation during tensile loading. In the thermal case, they predict increasing surface-feature size with temperature, whereas the reference shows an overall decrease. The checkpoint matrices reveal when later observations correct these errors and where they persist (Fig. 5e,h). Fable gains more from the observations, but both models retain low final numerical precision. Across twenty fixed-policy cases and 48 common numerical endpoints, case-equal scores rise from 32.00% to 37.50% for Astra and from 21.42% to 37.67% for Fable. Fable's larger gain closes its initial disadvantage without producing a higher final score. Absolute errors decrease on 16 Astra endpoints and 21 Fable endpoints, but increase on nine and fourteen. New observations therefore produce selective quantitative repair alongside harmful revisions (Supplementary Note 8, Supplementary Table 11, Supplementary Figs. 22–27).

Earlier measurements can substantially improve quantitative predictions of downstream material states. In continuously loaded laser-powder-bed-fused AlSi10Mg, Astra revises the hidden 5% strain dislocation density from $1.2 \times 10^{15}$ to $1.02 \times 10^{16}$ $m^{-2}$ after preceding density observations. Against the $9.5 \times 10^{15}$ $m^{-2}$ reference, relative error falls from 87.4% to 7.4% (Fig. 5f). The model successfully uses the preceding defect population to calibrate a later deformation state.

For $TiB_2$/AlSi10Mg undergoing solution treatment and ageing, Astra's tensile-strength prediction moves from 445 through 435 and 430 MPa before returning to 445 MPa. The reference is 431.3 MPa. Its error first falls from 13.7 to 1.3 MPa, then returns to 13.7 MPa as further structural evidence is incorporated. Additional evidence can also destabilize an initially accurate prediction when competing mechanisms must be combined. Astra temporarily reaches the experimental strength but loses this accuracy after further updates. This reveals a weakness in maintaining the quantitative balance among network degradation, solute redistribution and precipitation strengthening. The NbMoTaW-film case shows partial calibration followed by renewed error. For the hidden 625 °C surface-feature size, Astra changes from 360 to 100 and then 150 nm; Fable changes from 290 to 115 and then 140 nm. The reference is 51 nm (Fig. 5g,h). Both final estimates remain more than 2.7 times the reference, and both are worse than their earlier estimates. Additional observations correct part of the initial overestimate but fail to establish a stable quantitative description of the thermal response [37,38].

Accurate route selection requires quantitative consistency between predicted properties and design constraints. Fable selects the reference T6 route while predicting yield and tensile strengths below the stated thresholds, and acknowledges the shortfalls. The model identifies the reference route but cannot justify its feasibility using its own quantitative predictions. This inconsistency limits the reliability of its materials-design judgement. The largest gap in ME20 lies between understanding released experimental records and inferring material responses that remain unknown. Both Astra and Fable recover all released records, reaching 100% across the twenty cases (Fig. 5i). However, their scores for hidden forecasts remain at 49.2% and 46.0%, respectively. Even at the last hidden endpoints, the scores reach only 51.0% for Astra and 58.0% for Fable. These results reveal a substantial gap between assimilating observed data and using it to infer unresolved material states. Progressive evidence provides partial improvements, but the models still struggle to maintain quantitatively reliable judgements across extended research pathways. This limitation is critical for laboratory reasoning systems, where each new measurement must inform the interpretation of subsequent states, mechanisms and properties (Supplementary Table 12, Supplementary Fig. 28).

Prediction gains are uneven across material properties, even when models receive the same sequence of multiscale observations (Supplementary Fig. 29). Across the same 48 terminal numerical targets per model, Astra gains three and loses two top-band predictions. Fable gains four and loses one (Supplementary Fig. 30). Both models achieve a small net gain in top-band predictions, but some initially accurate responses become incorrect after updating. Endpoint gains alone therefore overestimate the reliability of sustained materials reasoning.

Together, M1, M2 and ME20 reveal three connected limitations in laboratory-facing materials reasoning: inaccurate condition-specific estimates, incomplete transfer of related knowledge and unstable predictions during evidence updates. M1 and M2 examine scientific judgements at individual research nodes, while ME20 tests how these judgements interact across processing histories, structural scales and sequential observations. Their combined results show that reliable scientific intelligence requires quantitative accuracy to be established, transferred and maintained throughout an evolving investigation.

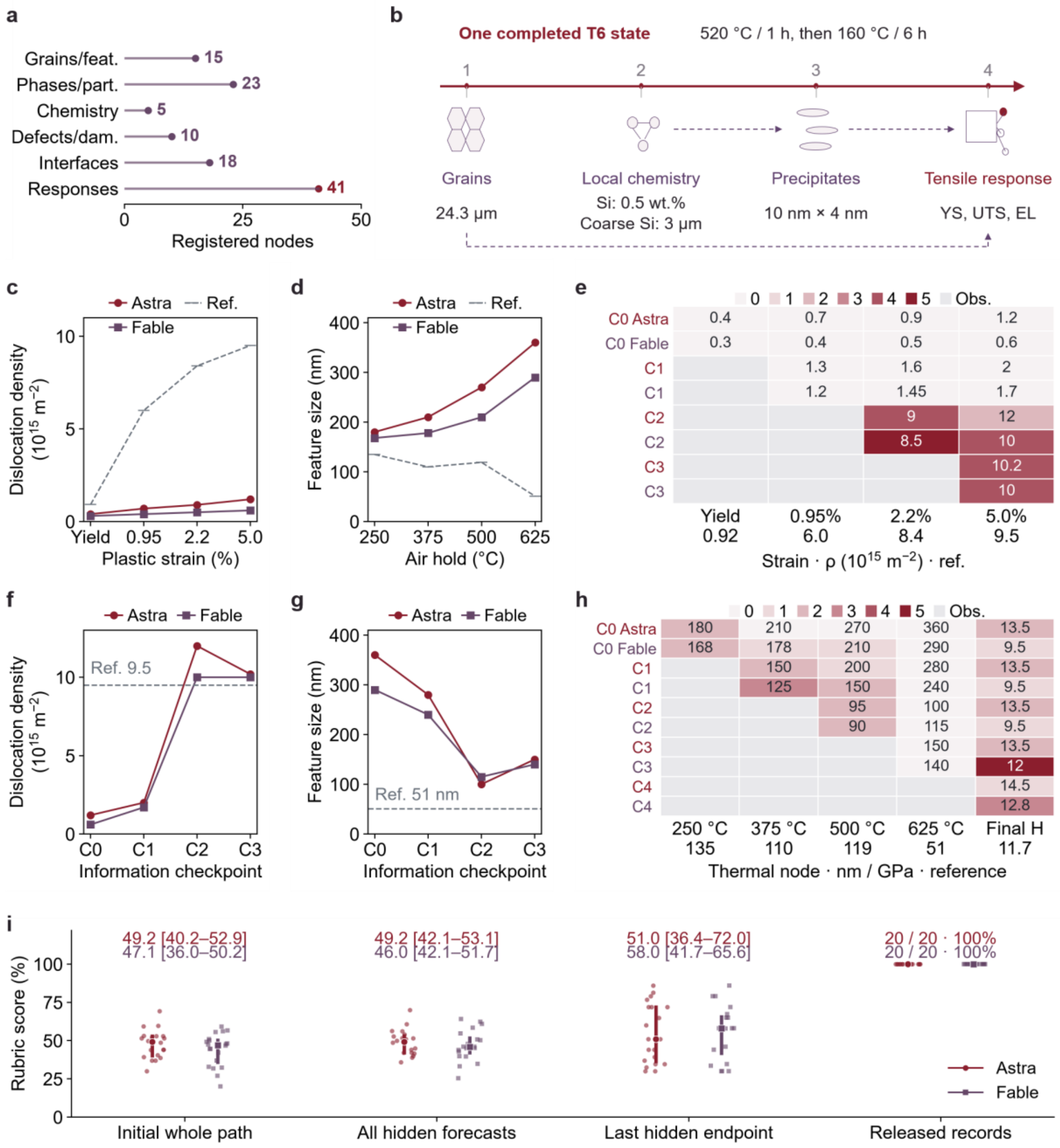

**Fig. 5 | Whole-path material prediction and sequential information-driven updating. a,** Ninety registered nodes across twenty cases, classified into six non-exclusive material-observable families. **b,** Four characterization records from one completed T6 state. Solid lines connect the shared processing state to its measurements. Dashed links show registered cross-scale constraints. Numbered measurements follow the information-release order. Icons illustrate observable families. Panels are read from left to right within each row: c, d, e above f, g, h. The deformation panels c and f form one column, the thermal panels d and g another, and the checkpoint matrices e and h the third. **c,** Initial forecasts of Al-matrix dislocation density across four plastic-strain states. The horizontal axis is strain state and the vertical axis density in $10^{15}$ $m^{-2}$. **d,** Initial forecasts of FESEM surface-feature size at successive 2-h air holds, with temperature on the horizontal axis and size in nm on the vertical axis. c and d compare two deployments with observation references. **e,** Predictions at every information checkpoint for the deformation case. Columns identify target states, rows checkpoint and deployment. Cell text gives the submitted value, colour the numerical grade, and grey cells already released results. **f,** Revision of the 5% strain density endpoint before its release. Astra changes from $1.2 \times 10^{15}$ to $1.02 \times 10^{16}$ $m^{-2}$, against $9.5 \times 10^{15}$ $m^{-2}$, increasing grade 0/5 to 4/5. **g,** Revision of the hidden 625 °C surface-feature endpoint, with an observed reference of 51 nm. **h,** Predictions at every information checkpoint for the thermal case, using the same cell conventions as e. Feature size in nm and hardness in GPa retain their separate units. **i,** Score distributions across the same twenty cases per deployment, shown on one shared axis in four categories: initial full-path forecast, unrevealed forecasts, last hidden endpoint and released-record recovery. Points are cases, thick lines span the interquartile range, and labels give median and quartiles. The released-record recovery category retains 40/40 model–case pairs at 100%; these responses reproduce records already available at submission.

## Discussion

MatScope reveals a fundamental gap between general scientific competence and the state-specific quantitative reasoning required for self-evolving materials laboratories. Across leading LLMs, strong performance in established calculations and physical explanations does not translate into reliable discrimination of material states, effective transfer of related knowledge, or stable prediction updates as experimental evidence accumulates. These findings identify a central requirement for scientific intelligence: models must establish accurate material-specific judgements, transfer relevant knowledge across conditions, and maintain quantitative consistency throughout evolving investigations. By connecting local reasoning, knowledge acquisition and sequential evidence updates, MatScope provides a unified evaluation framework for materials research functions beyond the alloy systems examined here. It establishes actionable targets for developing and evaluating scientific agents, materials-aware world models and self-evolving laboratory systems.

## Methods

### Corpus, task construction and information control

The analyses use the English Core200 questions, twenty paired M2 cases and twenty ME20 pathways. The corpus contains 28,872 deduplicated works and 716,080 indexed text passages. Structured extractions index candidate measurements and scientific relationships. Source text and structured measurements provide the reference values and relationships. Each subquestion is assigned a scientific purpose, a source-supported reference and a target-specific scoring rule. The question specifies the relevant composition, processing history, specimen state and measurement definition. Calculation targets include the required operands. Empirical prediction targets ask for a response under the stated conditions. Comparisons specify their reference direction, and decision targets specify their constraints. Samples, processing branches, states, locations, physical quantities and statistical definitions remain associated. Original observations, calculations and author interpretations are distinguished. Metadata, embeddings and UMAP characterize coverage. Historical task refinement informed the final development set, so its composition is reported explicitly. Current model-facing experiments use text and numerical records of characterization [39,40].

M1 questions provide the material conditions and answer requirements without tools or retrieved sources. Each task begins in an independent context with fixed model weights. The evaluation retains model, provider, generation settings, requests, submitted answers and termination states. Each M1 task starts in an independent conversation. The model receives the English question and its answer requirements. Tools and retrieved documents are disabled. Only the submitted final answer is scored. Returned reasoning and operational records are retained for diagnostic analysis.

M2 evaluates twenty selected problems from M1 nineteen source-connected groups. The paired comparison uses the same 64 subquestions and 66 targets in M1 and M2. Each episode starts without the earlier M1 answer, score or grader feedback. M2 adds the registered literature-search, reading and calculation tools within isolated episodes. The agent selects a search, passage-reading, calculation or submission action. The client executes the action and returns its observation to the same conversation. Each episode permits three searches, eight actions excluding submission, ten assistant turns and 6,000 observation tokens. The generated-token ceiling is 65,536, with 4,096 tokens reserved for the final answer. The episode ends at final submission or the 30-minute limit. Search combines BM25 and dense retrieval. Each initially supplies fifty candidates. Eligible candidates are combined by reciprocal-rank fusion and reranked with a Qwen model. Each search returns up to five permitted passages, capped at 500 tokens each. Additional reads return up to 800 tokens. After construction-source exclusion, retrieval searches and ranks candidates from 25,325 works and 712,778 chunks. Before text reaches the agent, a separate permission filter limits returned passages to 1,972 works and 63,639 chunks. Search coverage and readable body text therefore have different sizes. The agent chooses queries, reads current or adjacent passages, calculates and submits through tool calls (Supplementary Note 6, Supplementary Table 7). Twelve cases retain full-question inputs and eight use selected-question projections. Comparisons retain the same selected targets while recording this input-scope difference [41].

**Target-specific rubrics and experimental variability**

Each target specifies its scientific object, reference, accepted equivalences, required relationships, weight and scoring policy. Submitted final answers supply the scored claims. Scoring starts from the saved final answer. Reference-blind extraction identifies the submitted value, unit, object and state. Python checks numerical errors, unit conversions, identities and ordering. Gemini 3.1 Flash Lite evaluates the required explanatory relationships and their object and state assignments. The effective target grades are then combined using the task's registered weights and dependency rules. The range is 0–5. The highest grade requires all target-specific scientific requirements to be satisfied. Correct terminology alone does not satisfy a missing quantity, comparison or required relationship. Independent numerical and explanatory targets retain their separate grades. General statements earn credit only where they satisfy a stated scientific requirement. Contradictions and missing necessary relationships constrain the corresponding target. Joint M1 diagnostics require every prediction target to reach grade 5 and every mechanism target to reach at least grade 4. Verified extraction and scoring errors are corrected using the saved final answers. Remaining judgement concerns and their score sensitivity are reported separately (Supplementary Note 9, Supplementary Table 2) [42-44 45].

Numerical targets use absolute, relative or logarithmic-ratio errors according to the physical quantity. Signed differences retain their reference direction. Fraction errors preserve the distinction between percentage points and relative percentages. Empirical predictions receive graded credit under declared resolution bands. Deterministic calculations use calculation and rounding precision. Source-reported uncertainty and measurement definitions are retained where available. Prediction grades measure proximity to the empirical response at a declared resolution. Each reference retains its material state, measurement definition and reported statistic. Source-reported dispersion is recorded separately from the scoring bands. Raw errors, response orderings and constraint margins provide complementary measures of the scientific consequences of an error. Raw errors, orderings and constraint margins supplement rubric scores so that a near-threshold miss can be distinguished from an incorrect scale or decision.

Within a task, applicable target grades are combined using specified normalized weights and declared dependency rules. The task score has a 20% ceiling when none of its scored claims meets the rubric's criterion for a material-specific discriminating response. Mean Rubric Score averages task scores. Task Success follows the rule specified for each task, including designated target thresholds and, where specified, terminal research-goal conditions. Target-type analyses retain their own denominators. Missing final answers enter the end-to-end denominator as zero and receive a separate operational label. Completed-answer analyses are reported separately. Identified extraction, unit, ordering and semantic discrepancies are documented in the evaluation records, with their effect on scores reported separately. The question set removes two targets with an unspecified local sampling object or ledger operand, consecutively renumbers the retained subquestions, and replays saved answers under the shared unit and ordering repairs. Original requests, answers and scores are retained.

**Sequential forecasting and prediction stability**

ME20 records the process, material-state, measurement and information-release graphs separately. A scientific node specifies a material state, measurement object and requested response. A checkpoint specifies when the model submits predictions and receives an observation. Seventy nodes concern hidden measurements of an existing state, and twenty concern later physical states. Material relationships determine which hidden targets may be informed by each observation. The initial response predicts all requested hidden targets before any observation is disclosed. Temporal evolution, shared-state measurements and cross-scale links connect earlier nodes to relevant later quantities. Parallel measurements can constrain a farther endpoint or a joint choice while leaving the next-numbered node unchanged (Supplementary Note 7, Supplementary Table 10). At each checkpoint, the model predicts the targets that remain hidden. An explicit KEEP response refers to the same target at a named earlier checkpoint. Omitted targets remain missing. Values submitted after disclosure are excluded from hidden-response scores. Reproduction of released records is scored separately. Subsequent observations are then disclosed. Forecast scores exclude submissions made after target disclosure. Released-fact recovery is assessed separately. Each target is averaged over its required hidden checkpoints. Nodes combine numerical and relation components at the specified 70:30 weighting, followed by node- and case-level averaging. Initial and last-hidden endpoint comparisons retain the same targets. Design analysis distinguishes option identity, feasibility under the model's own numbers, and accuracy of the numerical support.

The uniform scoring snapshot contains 52 saved trajectories and 340 checkpoint finals, with all 52 whole-trajectory evaluations closed. Each deployment has 20 fixed-policy and six active-policy trajectories. The same specified judge configuration was retained for targeted completion of saved answers, without regenerating any scientific response. Whole-episode fixed-policy comparisons now use all twenty common cases. The 48 matched hidden numerical endpoints retain their separate twenty-case denominator. No historical mixed-configuration grade is borrowed.

Numerical repeatability was also examined on twenty preselected M1 tasks, using one target per task and three independent responses from Astra and Opus 5.5. Nineteen comparable triples were available per model. We divide the range of same-unit predictions by the target's declared top-band error scale. This measures response variability relative to evaluation resolution. It is distinct from experimental scatter and from accuracy against the reference. Statistical intervals resample connected source groups, with 2,000 replicates for the current M1, M2 and ME20 summaries. These intervals describe task-source variation. Repeat-generation results describe model variability. Each analysis reports its data source, sample size and scoring criteria (Supplementary Note 5, Supplementary Table 6, Supplementary Figs. 12–14) [46].

## Data availability

The benchmark questions, reference answers, and scoring rubrics are provided in the supplementary files submitted with this manuscript and will be made available at https://github.com/dddxs/MatScope upon publication. All relevant data supporting the findings of this study are included in the article and Supplementary Information.

## Code availability

The code used in this study will be made available to editors and reviewers upon request during peer review and is planned for public release at https://github.com/dddxs/MatScope upon publication.

## References

1 Ghafarollahi, A. & Buehler, M. J. Automating alloy design and discovery with physics-aware multimodal multiagent AI. *Proc. Natl Acad. Sci. USA* **122**, e2414074122 (2025).

2 Boiko, D. A., MacKnight, R., Kline, B. & Gomes, G. Autonomous chemical research with large language models. *Nature* **624**, 570-578 (2023).

3 Polak, M. P. & Morgan, D. Extracting accurate materials data from research papers with conversational language models and prompt engineering. *Nat. Commun.* **15**, 1569 (2024).

4 Kang, Y. & Kim, J. ChatMOF: an artificial intelligence system for predicting and generating metal-organic frameworks using large language models. *Nat. Commun.* **15**, 4705 (2024).

5 Bajan, C. & Lambard, G. Exploring the expertise of large language models in materials science and metallurgical engineering. *Digital Discovery* **4**, 500-512 (2025).

6 Zhang, J. et al. MatSciBench: Benchmarking the Reasoning Ability of Large Language Models in Materials Science. Preprint at https://arxiv.org/abs/2510.12171 (2025).

7 Alampara, N. et al. Probing the limitations of multimodal language models for chemistry and materials research. *Nature Computational Science* **5**, 952-961 (2025).

8 Xie, H. et al. LightAlloy-Bench: Benchmarking reasoning abilities of large language models in light alloys. *Journal of Magnesium and Alloys* **22**, 102127 (2026).

9 Yoshitake, M., Suzuki, Y., Igarashi, R., Ushiku, Y. & Nagato, K. MaterialBENCH: Evaluating College-Level Materials Science Problem-Solving Abilities of Large Language Models. Preprint at https://arxiv.org/abs/2409.03161 (2024).

10 Cheung, J. J. et al. MSQA: Benchmarking LLMs on Graduate-Level Materials Science Reasoning and Knowledge. Preprint at https://arxiv.org/abs/2505.23982 (2025).

11 Niyongabo Rubungo, A., Li, K., Hattrick-Simpers, J. & Bousso Dieng, A. LLM4Mat-bench: benchmarking large language models for materials property prediction. *Machine Learning: Science and Technology* **6**, 020501 (2025).

12 Mirza, A. et al. A framework for evaluating the chemical knowledge and reasoning abilities of large language models against the expertise of chemists. *Nature Chemistry* **17**, 1027-1034 (2025).

13 Kim, H. et al. Towards Fully-Automated Materials Discovery via Large-Scale Synthesis Dataset and Expert-Level LLM-as-a-Judge. Preprint at https://arxiv.org/abs/2502.16457 (2025).

14 Wang, X. et al. SciBench: evaluating college-level scientific problem-solving abilities of large language models. In *41st International Conference on Machine Learning* **235**, 50622-50649 (PMLR, 2024).

15 Rein, D. et al. GPQA: A Graduate-Level Google-Proof Q&A Benchmark. In *First Conference on Language Modeling* (2024).

16 Bragg, J. et al. AstaBench: Rigorous Benchmarking of AI Agents with a Scientific Research Suite. In *International Conference on Learning Representations* (2026).

17 Shen, Y. et al. SciAgentGym: Benchmarking Multi-Step Scientific Tool-Use in LLM Agents. In *43rd International Conference on Machine Learning* **306**, 111430-111454 (PMLR, 2026).

18 Mandal, I. et al. Evaluating large language model agents for automation of atomic force microscopy. *Nat. Commun.* **16**, 9104 (2025).

19 Liu, S. et al. MatTools: Benchmarking Large Language Models for Materials Science Tools. Preprint at https://arxiv.org/abs/2505.10852 (2025).

20 Liu, X. et al. AgentBench: Evaluating LLMs as Agents. In *International Conference on Learning Representations* (2024).

21 Mialon, G. et al. GAIA: A Benchmark for General AI Assistants. In *International Conference on Learning Representations* (2024).

22 Chen, Z. et al. ScienceAgentBench: Toward Rigorous Assessment of Language Agents for Data-Driven Scientific Discovery. In *International Conference on Learning Representations* (2025).

23 Laurent, J. M. et al. LAB-Bench: Measuring Capabilities of Language Models for Biology Research. Preprint at https://arxiv.org/abs/2407.10362 (2024).

24 Padovano, E., Badini, C., Pantarelli, A., Gili, F. & D'Aiuto, F. A comparative study of the effects of thermal treatments on AlSi10Mg produced by laser powder bed fusion. *J. Alloys Compd.* **831**, 154822 (2020).

25 Tan, Q. et al. A novel method to 3D-print fine-grained AlSi10Mg alloy with isotropic properties via inoculation with LaB6 nanoparticles. *Additive Manufacturing* **32**, 101034 (2020).

26 Cai, Q., Mendis, C. L., Chang, I. T. H. & Fan, Z. Microstructure evolution and mechanical properties of new die-cast Al-Si-Mg-Mn alloys. *Materials & Design* **187**, 108394 (2020).

27 McInnes, L., Healy, J. & Melville, J. UMAP: Uniform Manifold Approximation and Projection for Dimension Reduction. Preprint at https://arxiv.org/abs/1802.03426 (2018).

28 Ondicho, I. et al. Effect of phase reversion and recrystallisation on the microstructure and mechanical properties of a metastable medium-entropy alloy. *Mater. Sci. Eng. A* **949**, 149344 (2026).

29 Rödler, G. et al. Additive manufacturing of high-strength eutectic aluminium-nickel alloys – Processing and mechanical properties. *Journal of Materials Processing Technology* **298**, 117315 (2021).

30 Bogdanoff, T., Lattanzi, L., Merlin, M., Ghassemali, E. & Seifeddine, S. The influence of copper addition on crack initiation and propagation in an Al–Si–Mg alloy during cyclic testing. *Materialia* **12**, 100787 (2020).

31 Fields, B. et al. Investigation of an additively manufactured modified aluminum 7068 alloy: Processing, microstructure, and mechanical properties. *Mater. Sci. Eng. A* **891**, 145901 (2024).

32 Pourgharibshahi, M., Nobari, H. & Raiszadeh, R. An investigation on effect of tramp elements and solidification processing on homogenization kinetics of AA 7xxx series aluminum alloys. *Journal of Materials Research and Technology* **33**, 4674-4684 (2024).

33 Friel, R., Belyi, M. & Sanyal, A. RAGBench: Explainable Benchmark for Retrieval-Augmented Generation Systems. Preprint at https://arxiv.org/abs/2407.11005 (2024).

34 Niu, C. et al. RAGTruth: A Hallucination Corpus for Developing Trustworthy Retrieval-Augmented Language Models. In *62nd Annual Meeting of the Association for Computational Linguistics*, 10862-10878 (Association for Computational Linguistics, 2024).

35 Bolzoni, L., Nowak, M. & Hari Babu, N. Grain refinement of Al–Si alloys by Nb–B inoculation. Part II: Application to commercial alloys. *Materials & Design* **66**, 376-383 (2015).

36 Merino, J. et al. Multiple, comparative heat treatment and aging schedules for controlling the microstructures and mechanical properties of laser powder bed fusion fabricated AlSi10Mg alloy. *Journal of Materials Research and Technology* **13**, 669-685 (2021).

37 Zhang, X. X. et al. Evolution of microscopic strains, stresses, and dislocation density during in-situ tensile loading of additively manufactured AlSi10Mg alloy. *International Journal of Plasticity* **139**, 102946 (2021).

38 Wang, A. et al. Simultaneously enhancing strength and ductility of laser directed energy deposited TiB2/AlSi10Mg by inducing dispersed Mg2Si precipitation and reducing strain concentration. *International Journal of Plasticity* **193**, 104455 (2025).

39 Sutton, C. et al. Identifying domains of applicability of machine learning models for materials science. *Nat. Commun.* **11**, 4428 (2020).

40 Liang, P. et al. Holistic Evaluation of Language Models. Preprint at https://arxiv.org/abs/2211.09110 (2022).

41 Robertson, S. & Zaragoza, H. The Probabilistic Relevance Framework: BM25 and Beyond. *Foundations and Trends® in Information Retrieval* **4**, 1-174 (2009).

42 Starace, G. et al. PaperBench: Evaluating AI's Ability to Replicate AI Research. In *Proceedings of the 42nd International Conference on Machine Learning* **267** (PMLR, 2025).

43 Zheng, L. et al. Judging LLM-as-a-Judge with MT-Bench and Chatbot Arena. In *37th Conference on Neural Information Processing Systems* **36** (2023).

44 Liu, Y. et al. G-Eval: NLG Evaluation using GPT-4 with Better Human Alignment. In *Conference on Empirical Methods in Natural Language Processing*, 2511-2522 (Association for Computational Linguistics, 2023).

45 Kim, S. et al. Prometheus 2: An Open Source Language Model Specialized in Evaluating Other Language Models. In *Proceedings of the 2024 Conference on Empirical Methods in Natural Language Processing*, 4334-4353 (Association for Computational Linguistics, 2024).

46 Efron, B. Bootstrap Methods: Another Look at the Jackknife. *The Annals of Statistics* **7** (1979).

**Acknowledgments**

X.D. acknowledges financial support from the China Scholarship Council (CSC). We thank Songmao Liang from CompuTherm LLC for the Pandat technique support. During the preparation of this work, we used the LLM ChatGPT by OpenAI to refine the language and enhance the readability of this study. After using this tool, we reviewed and edited the content as needed and take full responsibility for the content of the publication.

**Funding**

This work was supported by the Science and Technology Cooperation Program of Shanghai Jiao Tong University in the Inner Mongolia Autonomous Region—Action Plan of Shanghai Jiao Tong University for "Revitalizing Inner Mongolia through Science and Technology" (grant no. 2023XYJG0001-01-01) and the National Research Foundation, Singapore, under its AI Singapore Programme (AISG Award No. AISG2-GC-2023-009).

**Competing interests:** The authors declare that they have no known competing financial interests or personal relationships that could have appeared to influence the work reported in this paper.

**Author information**

**Authors and Affiliations**

**Shanghai Key Lab of Advanced High-temperature Materials and Precision Forming and State Key Lab of Metal Matrix Composites, School of Materials Science and Engineering, Shanghai Jiao Tong University, Shanghai, People's Republic of China**

Xuyu Dong, Da Shu

**School of Materials Science and Engineering, Nanyang Technological University, Singapore, Singapore**

Xuyu Dong, Zheng Liu

**College of Computing and Data Science, Nanyang Technological University, Singapore, Singapore**

Xuyu Dong, Yixuan Li, Jianguo Huang, Feng Chen, Yiyun Zhou, Bo An

**Author contributions**

Conceptualization: X.D., D.S., B.A.
Experiments: X.D.,
Data engineering: X.D., J.H.
Framework design and construction: X.D.
Analysis: X.D.,
Visualization: X.D., Y.L., F.C., Y.Z.
Supervision: X.D., D.S., Z.L., B.A.
Funding: D.S., Z.L., B.A.
Writing—original draft: X.D.
Writing—review & editing: X.D., D.S., Z.L.

**Corresponding authors**

Correspondence to Bo An or Zheng Liu or Da Shu.